\documentclass[lettersize,journal]{IEEEtran}
\usepackage{amsmath,amsfonts}
\usepackage{algorithmic}
\usepackage{algorithm}
\usepackage{array}
\usepackage{bm}
\usepackage{booktabs}
\usepackage[caption=false,font=normalsize,labelfont=sf,textfont=sf]{subfig}
\usepackage{multirow}
\usepackage{textcomp}
\usepackage{stfloats}
\usepackage{url}
\usepackage{verbatim}
\usepackage{graphicx}
\usepackage{cite}
\begin{document}

\title{Content-Adaptive Image Retouching Guided by Attribute-Based Text Representation}

\author{Hancheng Zhu,
        Xinyu Liu,
        Rui Yao,
        Kunyang Sun,
        Leida Li,
        and Abdulmotaleb El Saddik,~\IEEEmembership{Fellow,~IEEE}
\thanks{This work was supported in part by the National Natural Science Foundation of China under Grants 62101555, 62172417, and 62471349, in part by the Natural Science Foundation of Jiangsu Province under Grant BK20210488.}
\thanks{H. Zhu, X. Liu, R. Yao and K. Sun are with the School of Computer Science and Technology/School of Artificial Intelligence, China University of Mining and Technology, Xuzhou 221116, China, and also with the Mine Digitization Engineering Research Center of the Ministry of Education, China University of Mining and Technology, Xuzhou, 221116, China (e-mails: zhuhancheng@cumt.edu.cn; xinyu\_liu\_cs@163.com; ruiyao@cumt.edu.cn; kunyang\_sun@cumt.edu.cn).}
\thanks{L. Li is with the School of Artificial Intelligence, Xidian University, Xi'an 710071, China (e-mail: ldli@xidian.edu.cn).}
\thanks{Abdulmotaleb El Saddik is with the School of Electrical Engineering and Computer Science, University of Ottawa, Ottawa, ON K1N 6N5, Canada (e-mail: elsaddik@uottawa.ca)}
}

\markboth{Journal of \LaTeX\ Class Files,~Vol.~14, No.~8, August~2021}%
{Shell \MakeLowercase{\textit{et al.}}: A Sample Article Using IEEEtran.cls for IEEE Journals}


\maketitle

\begin{abstract}
Image retouching has received significant attention due to its ability to achieve high-quality visual content. Existing approaches mainly rely on uniform pixel-wise color mapping across entire images, neglecting the inherent color variations induced by image content. This limitation hinders existing approaches from achieving adaptive retouching that accommodates both diverse color distributions and user-defined style preferences. To address these challenges, we propose a novel Content-Adaptive image retouching method guided by Attribute-based Text Representation (CA-ATP). Specifically, we propose a content-adaptive curve mapping module, which leverages a series of basis curves to establish multiple color mapping relationships and learns the corresponding weight maps, enabling content-aware color adjustments. The proposed module can capture color diversity within the image content, allowing similar color values to receive distinct transformations based on their spatial context. In addition, we propose an attribute text prediction module that generates text representations from multiple image attributes, which explicitly represent user-defined style preferences. These attribute-based text representations are subsequently integrated with visual features via a multimodal model, providing user-friendly guidance for image retouching. Extensive experiments on several public datasets demonstrate that our method achieves state-of-the-art performance.
\end{abstract}


\section{Introduction}
\IEEEPARstart{T}{he} ‌prevalence of smartphones and social networks has made images a dominant medium for life documentation and information sharing. This proliferation has consequently raised public expectations for image quality. However, raw photos often exhibit quality degradation due to poor lighting, adverse weather conditions, or limited photography skills. Although conventional image retouching methods can improve quality, their dependence on professional expertise and manual adjustments makes them impractical for most users. In recent years, advances in deep learning have spurred the development of automatic image retouching methods \cite{bychkovsky2011learning, chen2017fast}. These methods represent user preferences by simulating photography skills to obtain high-quality images, significantly promoting the widespread adoption of image retouching technology.

\begin{figure}[t]
\centering
\includegraphics[width=0.99\columnwidth]{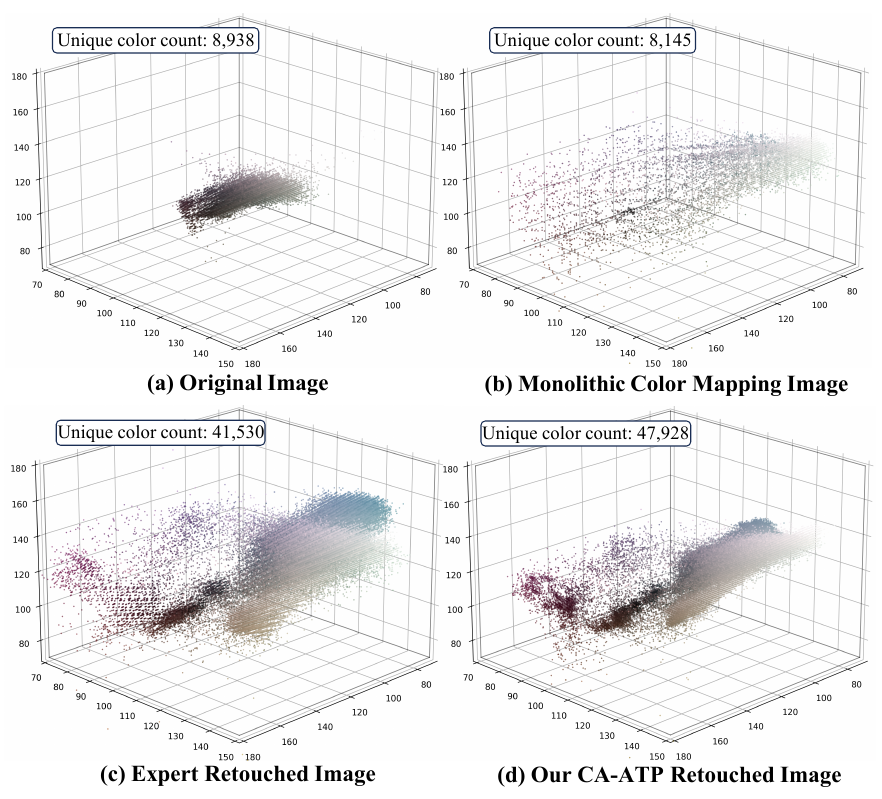} 
\caption{A case study of RGB color distributions (range 0-255) in the MIT-Adobe 5K dataset \cite{bychkovsky2011learning}. The ``unique color count" means the number of distinct RGB values, with each scatter point representing a unique color. The four 3D scatter plots (original image, monolithic color mapping image, expert retouched image, and our CA-ATP retouched image) show their distinct RGB space distributions.}
\label{fig1}
\end{figure}

Recent deep learning-based image retouching methods predominantly adopt a pixel-wise color mapping paradigm \cite{guo2020zero}. These approaches typically employ neural networks to learn global transformations (such as tone curves \cite{moran2021curl} or 3D lookup tables (3D LUTs) \cite{zeng2020learning}), which can establish mapping relationships through the RGB color space. However, such monolithic color mapping strategies \cite{kim2020global} fundamentally depart from human-like retouching workflows that integrate content-aware adjustments. These mappings demonstrate non-injective properties \cite{rosen1999discrete}, inherently limiting the color diversity  of the retouched images (measured by unique color count) to the range present in the original images. To demonstrate this limitation, we conduct a case study using the MIT-Adobe 5K dataset \cite{bychkovsky2011learning}, where comparative color distributions are presented in Fig.~\ref{fig1}. The results show that applying a monolithic color mapping model reduces the unique color count from approximately 9,000 in an original image (Fig.~\ref{fig1}(a)) to 8,145 (Fig.~\ref{fig1}(b)). In contrast, expert retouching significantly expands the color palette to 41,000 unique colors (Fig.~\ref{fig1}(c)), highlighting the color diversity of retouched images in practical situations. Hence, existing methods are constrained in the range of color adjustments, resulting in retouching results that fail to encompass image content with rich color information.

In addition, existing methods also exhibit limitations in representing user-defined style preferences \cite{kim2020pienet, bianco2020personalized}. Most methods typically leverage style vectors to represent different retouching styles~\cite{song2021starenhancer}. However, the key constraint of these methods lies in the high-dimensional abstraction of feature representation, which fails to explicitly capture users' style preferences \cite{wang2023learning, duan2025diffretouch}. In addition, some methods require additional user information when acquiring style vectors, which complicates image retouching in scenarios where user style preferences cannot be directly obtained.

To address these challenges, we introduce a content-adaptive image retouching method guided by attribute-based text representation, abbreviated as CA-ATP. Unlike conventional mapping strategies constrained by color diversity, CA-ATP emulates human content-aware adjustment ability. The proposed model consists of a U-Net \cite{ronneberger2015u} and a multimodal model \cite{radford2021learning}. The multimodal model generates a series of basis curves that include multiple color mapping relationships, while the U-Net dynamically learns content-aware weight maps. The proposed module enables pixel-level retouching through dynamic weighted combinations of all basis curves, wherein the weighting coefficients are content-adaptive rather than fixed. As a result, our method achieves content-aware color adjustments where perceptually similar colors receive differential processing. This mechanism fundamentally overcomes the intrinsic color space limitations of the original images, achieving results comparable to expert retouching (as shown in Fig.~\ref{fig1} (d), which demonstrates the retouching results of our CA-ATP method).

To intuitively and effectively obtain user style preferences, we propose an attribute-based text representation strategy. Specifically, we utilize quantifiable attributes (such as brightness, saturation, and contrast) to characterize images of different styles and construct semantically meaningful style attribute vectors. Then, these attribute vectors are translated into text descriptions, enabling a more explicit representation of user-defined style preferences. Based on user-retouched images, we develop an attribute text prediction module that can automatically generate attribute text prompts that align with user preferences without requiring additional information. Hence, our approach enables the control of image retouching direction through explicit attribute text prompts, which is a user-friendly image retouching strategy. In summary, our main contributions are as follows.

\begin{itemize}
\item We propose a content-aware curve mapping framework that adaptively blends a series of basis curves via content-aware weight maps, allowing content-adaptive adjustments of similar colors in an image.
\item We develop an attribute-based text representation strategy that can automatically learn explicit representations of user-defined style preferences from their retouched images, which is user-friendly for controlling the direction of image retouching.
\item Extensive experiments on multiple public benchmarks demonstrate that our approach achieves state-of-the-art performance in image retouching, validating its effectiveness and superiority.
\end{itemize}


\begin{figure*}[!t]
\centering
\includegraphics[width=0.95\textwidth]{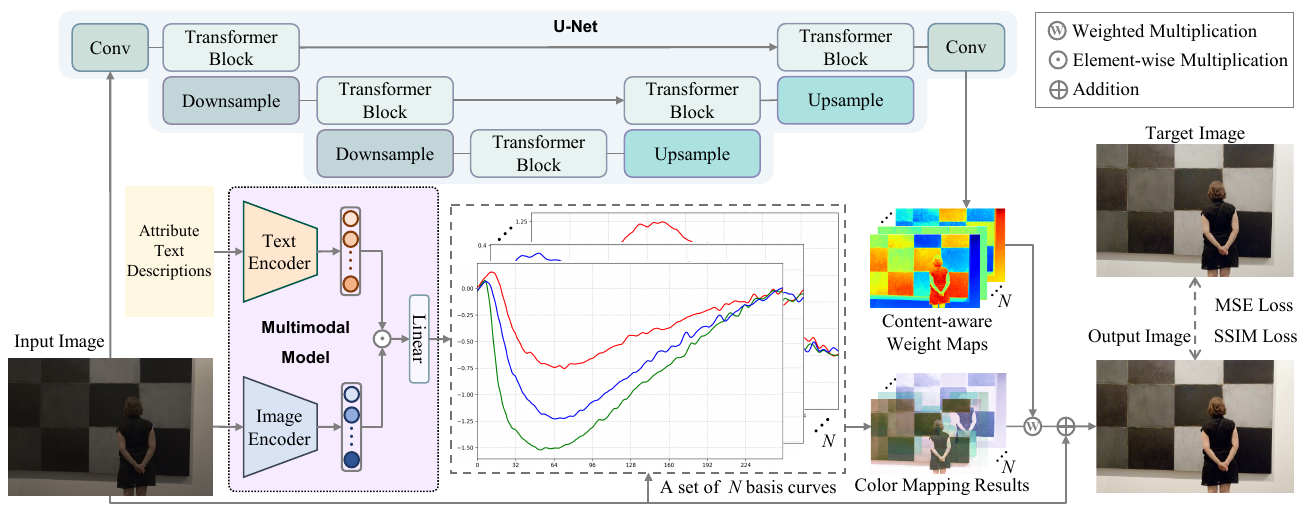}
\caption{The framework of our model comprises two parallel branches. The bottom branch utilizes a multimodal model, ‌performing multiplicative fusion on features extracted from input images and attribute text descriptions, ‌generating $N$ basis curves ‌to derive $N$ color mapping results. Concurrently, the top branch employs a U-Net ‌to produce $N$ pixel-level content-aware weight maps ‌based on input images. Finally, content-adaptive ‌results are generated ‌through weighted fusion of the $N$ color mapping results, ‌guided by these weight maps.}
\label{fig2}
\end{figure*}

\section{Related Works}

\subsection{Color Mapping-based Image Retouching}
Color mapping-based image retouching methods can be broadly categorized into curve-based and Lookup Table (LUT)-based techniques. Curve-based methods enhance images by mapping pixel values through tone adjustment curves. This mapping technique was traditionally used as a classical enhancement tool \cite{6213525, 8936990}. Recently, it has been integrated with deep learning, employing neural networks to predict the optimal curve directly. Zero-DCE \cite{guo2020zero} pioneered this paradigm by integrating deep curve estimation for image enhancement \cite{wei2018deep}. CURL \cite{moran2021curl} was proposed by cascading multiple curves across RGB, HSV, and LAB color spaces. To improve nonlinear adjustment capability, FlexiCurve \cite{li2023flexicurve} introduced piecewise differentiable curve designs. NamedCurves \cite{serrano2024namedcurves} decomposed images into named color regions \cite{berlin1991basic} and applied tone curve adjustments to each color category. LUT-based methods are widely adopted across various computer vision tasks, such as image pipelines, editing software \cite{karaimer2016software}, and watermarking \cite{1227614}. 3D LUT \cite{zeng2020learning} was the first to introduce 3D LUTs into deep learning frameworks. To enhance sampling flexibility, Yang \emph{et al.} proposed an adaptive interval sampling scheme for non-uniform sampling in 3D color spaces \cite{yang2022adaint}. They further proposed SepLUT \cite{yang2022seplut}, which decomposes a complex transformation into a cascade of 1D and 3D LUTs. Finally, 4DLUT \cite{liu20234d} extended the LUT to four dimensions, incorporating more spatial information. Although the above methods have achieved remarkable performance, they neglect the diverse mapping relationships of colors in image retouching. In contrast, our approach learns a broader spectrum of diversity relationships in color mapping through content-adaptive mechanisms.

\subsection{Multi-style Image Retouching}
Early image retouching approaches are limited to producing a single fixed style, which restricts their ability to satisfy diverse user preferences. To address this limitation, researchers have developed various style representations to empower a single model with multi-style image retouching. SpliNet \cite{bianco2020personalized} pioneered multi-style generation by using one-hot encoding to represent different styles. PieNet \cite{kim2020pienet} advanced this direction using embedded feature vectors for style representation. StarEnhancer \cite{song2021starenhancer} was proposed to incorporate a style transfer module \cite{huang2017arbitrary} and improve the flexibility of style control. Inspired by BERT \cite{devlin2019bert}, Kosugi \emph{et al.} introduced style masks, allowing the model to automatically infer the target style of an image \cite{kosugi2023personalized}. Kim \emph{et al.} further proposed multi-style control at the imaging pipeline level by predicting different image signal processor (ISP) parameters \cite{kim2023learning}. More recently, DiffRetouch \cite{duan2025diffretouch} was proposed to represent styles using numerical attributes and leverage contrastive learning \cite{henaff2020data} to improve multi-style capability. However, the aforementioned methods face two key issues. First, the abstract nature of style vectors prevents them from explicitly and intuitively representing user-defined style preferences. Second, some approaches require additional user information as a prerequisite for retouching, making it difficult to implement directly in inference \cite{duan2025diffretouch}. Diverging from existing approaches, we propose an attribute text prediction module that can automatically achieve user preference information. Moreover, the attribute text representation enables more intuitive and effective guidance for user-defined image retouching.

\section{Method}

\subsection{Overview}
\begin{figure}[t]
\centering
\includegraphics[width=0.85\columnwidth]{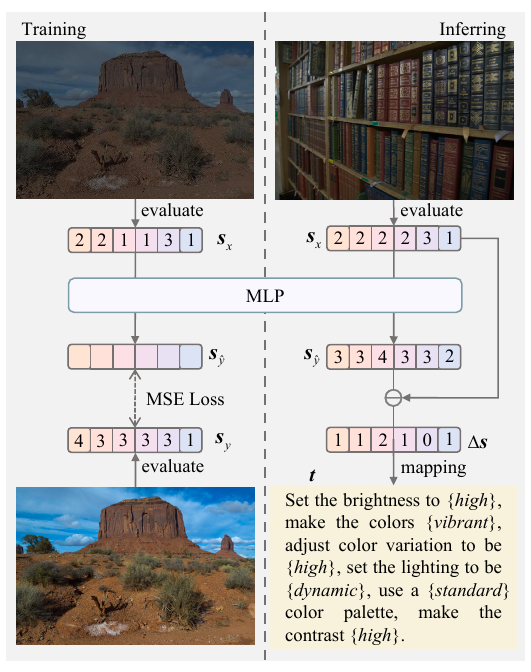}
\caption{The framework of the ATP module. Image styles are quantified through the attribute evaluation method, which maps complex visual characteristics into six intuitive numerical attributes. The six attributes are mean brightness, mean saturation, saturation standard deviation, brightness standard deviation, color richness, and contrast, and each attribute is discretized into 1 to 5 levels.}
\label{fig3}
\end{figure}

This section presents our proposed content-adaptive image retouching method guided by attribute-based text representation (CA-ATP), which consists of a content-adaptive curve mapping module and an attribute text prediction (ATP) module. As shown in Fig.~\ref{fig2}, our model contains two parallel branches: one employs a multimodal model to generate multiple sets of basis curves, while the other utilizes a U-Net architecture to produce the corresponding content-aware weight maps. To obtain text descriptions representing multiple attributes, the ATP module (as illustrated in Fig.~\ref{fig3}) can automatically generate text representations by learning the predictive attributes of each user's style preferences. This achieves explicit control over user-defined style preferences. Finally, diverse color mapping relationships for each pixel in images are achieved through a weighted combination of corresponding multiple sets of basis curves guided by content-aware weight maps. This enriches the color diversity of the images retouched by our CA-ATP model.

\subsection{Attribute Text Prediction Module}
To achieve an effective representation of users' diverse style preferences, we first quantify image styles using an attribute evaluation method. Different from previous methods (e.g., StarEnhancer \cite{song2021starenhancer}) that use abstract features extracted from deep networks, we map complex visual style characteristics to a set of intuitive numerical attributes. This process provides an objective and interpretable style representation. Based on the aesthetic attribute evaluation approach \cite{datta2006studying}, we employ six style-relevant attributes to quantify the style of images. The six attributes are mean brightness, mean saturation, saturation standard deviation, brightness standard deviation, color richness, and contrast, and each attribute is discretized into 1 to 5 levels. Therefore, the discretized attribute vector can be represented as $\bm{s} \in \mathbb{R}^6$. The mean values quantify the overall levels of these attributes, while the standard deviations reflect their distribution within the images.

A key aspect of multi-style image retouching is to learn a user's style preferences from previously retouched images and automatically apply these preferences to new images. Therefore, we introduce a Multilayer Perceptron (MLP) that predicts a user's preferred attribute vector. For a user, given training images $\bm{x}$ and its user-retouched counterpart $\bm{y}$, we use the attribute evaluation method \cite{datta2006studying} to obtain attribute vectors $\bm{s}_{x}$ and $\bm{s}_{y}$, respectively. We take $\bm{s}_{x}$ as input to predict the target attribute vector $\hat{\bm{s}}_{y}$ using $\mathcal{F}_\text{MLP}$:
\begin{equation}
\hat{\bm{s}}_{y} = \mathcal{F}_{\text{MLP}}(\bm{s}_{x}).
\end{equation}
During the training stage, the MLP is optimized by minimizing the mean squared error (MSE) loss between the predicted and user-retouched attribute vectors ($\hat{\bm{s}}_{y}$ and $\bm{s}_{y}$):
\begin{equation}
\mathcal{L}_{\text{MSE}} = \text{MSE}(\hat{\bm{s}}_{y}, \bm{s}_{y}).
\end{equation}

In the inference stage, given a new image, we first obtain its attribute vector $\bm{s}_{x}$ and leverage the trained MLP to predict the target attribute vector $\hat{\bm{s}}_{y}$. Then, we define the user preferred vector $\Delta\bm{s}$ as the difference between the predicted target and the input attribute vectors:
\begin{equation}
\Delta\bm{s} = \hat{\bm{s}}_{y} - \bm{s}_{x},
\end{equation}
where $\Delta\bm{s}$ indicates the desired direction and magnitude of attribute adjustments for image retouching. To convert these numerical attribute values into text descriptions, we design a text template-based mapping approach. Specifically, we pre-define a text template with placeholders for each attribute: \textit{``Set the brightness to \{level\}, make the colors \{level\}, adjust color variation to be \{level\}, set the lighting to be \{level\}, use a \{level\} color palette, make the contrast \{level\}.''} Each $\{level\}$ placeholder is determined based on the corresponding attribute value in the adjustment vector $\Delta\bm{s}$. For example, the five levels of brightness correspond to very low, low, medium, high, and very high. The specific descriptive terms for all attributes are listed in Table~\ref{table6}. More predefined description items and implementation details can be found in our source code.

Finally, the attribute text descriptions $\bm{t}$ can be ‌generated by populating the text template with terms corresponding to the user's preferred vector $\Delta\bm{s}$. This approach can ‌standardize text generation, which maintains user-friendliness in controlling image retouching directions while ensuring alignment with user-defined style preferences.

\begin{table*}[!t]
\caption{Mapping from numerical attribute values to predefined descriptive terms. Each of the six attributes is discretized into five levels. These descriptive terms are then inserted into the corresponding $\{level\}$ placeholders in our text template to generate attribute text descriptions.}
\label{table6}
\renewcommand{\arraystretch}{1.2}
\fontsize{8.5pt}{8.5pt}\selectfont%
\centering
\label{tab:attribute_mapping}
\begin{tabular}{cccccc}
\toprule
\textbf{Attributes} & $\Delta\bm{s}\ge1.5$ & $0.5\le \Delta\bm{s}< 1.5$ & $-0.5\le \Delta\bm{s}< 0.5$ & $-1.5\le \Delta\bm{s} <-0.5$ & $\Delta\bm{s}<-1.5$ \\
\midrule
\textbf{mean brightness} & very high & high & medium & low & very low \\
\textbf{mean saturation} & intensely vibrant & vibrant & natural & muted & desaturated \\
\textbf{saturation standard deviation} & extreme & high & moderate & low & minimal \\
\textbf{brightness standard deviation} & dramatic & dynamic & balanced & soft & flat \\
\textbf{color palette} & full-spectrum & rich & standard & limited & monochromatic \\
\textbf{contrast} & very high & high & medium & low & very low \\
\bottomrule
\end{tabular}
\end{table*}

\subsection{Content-Adaptive Curve Mapping Module}

\textbf{Curve Generation via Multimodal model.}
To enable our image retouching guided by attribute-based text representation, we adopt the pretrained CLIP \cite{radford2021learning} as our multimodal model, which maps images and text into a shared high-dimensional semantic embedding space, establishing cross-modal alignment.

We assume the training set for a user is $\{\bm{x}_i,\bm{y}_i\}_{i=1}^K$, where $\bm{y}_i$ denotes the target image corresponding to the input image $\bm{x}_i$, and $K$ represents the number of training image pairs. For each input image $\bm{x}_i$, the trained ATP module (as shown in Fig.~\ref{fig3}) can emulate the user-defined style preferences to generate an attribute text description $\bm{t}_i$. For each pair, we employ the image encoder $\mathcal{E}_{\text{img}}$ and text encoder $\mathcal{E}_{\text{txt}}$ to extract features from both the image $\bm{x}_i$ and its corresponding attribute text description $\bm{t}_i$:
\begin{equation}
\bm{f}_{\bm{x}_i} = \mathcal{E}_{\text{img}}(\bm{x}_i), \quad \bm{f}_{\bm{t}_i} = \mathcal{E}_{\text{txt}}(\bm{t}_i).
\end{equation}
The visual features $\bm{f}_{\bm{x}_i}$ and textual features $\bm{f}_{\bm{t}_i}$ are fusion through element-wise multiplication ($\odot$) to produce a multimodal feature $\bm{f}_i$:
\begin{equation}
\bm{f}_i = \bm{f}_{\bm{x}_i} \odot \bm{f}_{\bm{t}_i}.
\end{equation}
The fused feature $\bm{f}_i$ is then processed by a linear layer $\mathcal{F}_{\text{FC}}$ to generate a curve parameter vector $\bm{p}_i$:
\begin{equation}
\bm{p}_i = \mathcal{F}_{\text{FC}}(\bm{f}_i),
\end{equation}
where $\bm{p}_i \in \mathbb{R}^{3 \times N \times P}$ includes $N$ sets of basis curves. Each set comprises three curves (one per RGB channel) with $P$ control points per curve ($P=64$ in our implementation). To form continuous curves mapping all 256 intensity levels of an 8-bit color channel, we up-sample the $P$ sparse control points via bicubic interpolation. This produces $N$ sets of dense basis curves $\bm{c}_i \in \mathbb{R}^{3 \times N \times L}$ ($L=256$):
\begin{equation}
\bm{c}_i = \text{BicubicInterp}(\bm{p}_i, L).
\end{equation}

Finally, each of the $N$ curve sets is applied to the image $\bm{x}_i$ for generating a candidate image, yielding a set of $N$ color mapping results $\{\hat{\bm{x}}_{i,j}\}_{j=1}^N$, which is defined as:
\begin{equation}
\hat{\bm{x}}_{i,j} = \mathcal{T}(\bm{x}_i, \bm{c}_{i,j}),
\end{equation}
where $\mathcal{T}$ represents the curve-based color adjustment operation. The $N$ color mapping results are then propagated to a subsequent module for the final content-adaptive fusion.

\noindent\textbf{Content-aware Color Mapping.}
To obtain pixel-wise weight maps, we implement a content-aware module based on the U-Net architecture \cite{ronneberger2015u}. The U-Net is selected due to its skip connections and multi-scale feature fusion capabilities, which facilitate robust performance with limited training data. To augment global context modeling, we incorporate the Restormer architecture \cite{zamir2022restormer}. This method replaces the CNN blocks of U-Net with Transformer blocks \cite{dosovitskiy2021image}, enabling effective long-range dependency modeling. The details of the U-Net framework can be found in this work \cite{zamir2022restormer}.

Specifically, the U-Net $\mathcal{F}_{\text{Unet}}$ can embed an input image $\bm{x}_i$ to predict $N$ content-aware weight maps simultaneously, which can be defined as:
\begin{equation}
(\bm{w}_{i,1}, \bm{w}_{i,2}, \ldots, \bm{w}_{i,N}) = \mathcal{F}_{\text{Unet}}(\bm{x}_i),
\end{equation}
where $\bm{w}_{i,j}$ indicates a weight map for the $j$-th color mapping result, and it has the same shape as the image $\bm{x}_i$. 

The final retouched image $\hat{\bm{y}}_i$ is synthesized by fusing the $N$ candidate color mapping results $\{\hat{\bm{x}}_{i,j}\}_{j=1}^N$ with their associated weight maps. To ensure spatially coherent blending, we first apply channel-wise Softmax normalization to these weight maps:
\begin{equation}
\hat{\bm{w}}_{i,j} = \frac{\exp(\bm{w}_{i,j})}{\sum_{j=1}^{N} \exp(\bm{w}_{i,j})},
\end{equation}
followed by a weighted summation with element-wise multiplication:
\begin{equation}
\hat{\bm{y}}_i = \sum_{j=1}^{N} \hat{\bm{w}}_{i,j} \odot \hat{\bm{x}}_{i,j}.
\end{equation}
This strategy enables pixel-wise selection of optimal color adjustments, producing results that jointly achieve enhanced color diversity and precise content-aware retouching.

\subsection{Training Strategy}
To enable our model to perform style-specific image retouching based on attribute text descriptions, we introduce a specialized training data sampling strategy. For each image, we first create a pool containing all versions retouched by different users. During training, for the user-specific training dataset $\{\bm{x}_i,\bm{y}_i\}_{i=1}^K$, we randomly select one of these user-retouched images $\bm{y}_i$ as the target and pair the corresponding input image $\bm{x}_i$ with its attribute text description $\bm{t}_i$ for model training. This design explicitly reinforces the model to learn the correlation between attribute text representations and user visual preferences, thereby enhancing its intuitiveness and effectiveness in representing diverse user style preferences.

Specifically, we employ a joint loss function composed of the Mean Squared Error (MSE) and the Structural Similarity Index Measure (SSIM) \cite{zhao2016loss}. The MSE loss quantifies the pixel-wise difference between the retouched image $\hat{y}_i$ predicted by our model and the target image $\bm{y}_i$ for the user:
\begin{equation}
\mathcal{L}_{\text{MSE}} = \frac{1}{K} \sum_{i=1}^{K} \text{MSE}(\hat{\bm{y}}_i, \bm{y}_i).
\end{equation}
The SSIM loss evaluates perceptual similarity by considering structural information, luminance, and contrast, which aligns more closely with human visual perception \cite{wang2004image}. It is defined as:
\begin{equation}
\mathcal{L}_{\text{SSIM}} = \frac{1}{K} \sum_{i=1}^{K} \left( 1 - \text{SSIM}(\hat{\bm{y}}_i, \bm{y}_i) \right).
\end{equation}
Since the SSIM index ranges from $[0, 1]$, with higher values indicating greater similarity, we use its complement as loss.

Our final objective function ($\mathcal{L}_{\text{total}}$) is a weighted sum of these two losses:
\begin{equation}
\mathcal{L}_{\text{total}} = \alpha \mathcal{L}_{\text{MSE}} + \beta \mathcal{L}_{\text{SSIM}},
\end{equation}
where $\alpha$ and $\beta$ are hyperparameters that balance the contribution of each loss term.

\section{Experiments}
\subsection{Settings}

\noindent\textbf{Datasets.}
We evaluate our method on two standard datasets: MIT-Adobe 5K (MIT5K) \cite{bychkovsky2011learning} and PPR10K \cite{liang2021ppr10k}. ‌MIT5K comprises 5,000 RAW‌-format‌ images, ‌each retouched by five experts (A, B, C, D, and E) to provide five different reference styles. ‌PPR10K contains 11,161 portrait images organized into 1,681 groups, where every image is independently retouched by three experts (A, B, and C), yielding three references. 

We treat each expert as a user for image retouching. Image retouching can be divided into two principal categories: single-style methods for general users and multi-style approaches for the style preferences of multiple users. While our method focuses on multi-style image retouching, we also evaluate its performance in single-style image retouching scenarios. To ‌verify the effectiveness of our method, we compared it with other representative methods on these two paradigms.

\noindent\textbf{Single-Style Setup.}
Following the setup of UPE \cite{wang2019underexposed}, we use the first 4,500 images of MIT5K for training and the remaining 500 for testing. The original images are generated by decreasing the white balance of expert C’s version by 1.5 EV, and the targets are images retouched by expert C. All images are resized so that their shorter side is 512 pixels. This configuration is referred to as MIT5K-UPE. The performance of our model is evaluated using PSNR, SSIM, Learned Perceptual Image Patch Similarity (LPIPS) \cite{zhang2018unreasonable}, and the color difference $\Delta E_{ab}$.

\begin{table}[h]
\caption{
Peer comparison on the MIT5K-UPE dataset, where `-' represents unreported results. The best result is indicated in \textbf{bold}, while the second best is \underline{underlined}. The experimental setup follows the single-style setup described in Section 4.1. Some results are cited from the NamedCurves \cite{serrano2024namedcurves}.
}
\label{table1}
\centering
\renewcommand{\arraystretch}{1.2}
\fontsize{8.5pt}{8.5pt}\selectfont%
\begin{tabular}{ccccc}
\toprule
\textbf{Method} & \textbf{PSNR}$\uparrow$ & \textbf{SSIM}$\uparrow$ & \textbf{LPIPS}$\downarrow$ & $\boldsymbol{\Delta E_{ab}}$$\downarrow$ \\
\midrule
UPE \cite{wang2019underexposed}          & 23.04 & 0.893 & 0.158 & --   \\
CURL \cite{moran2021curl}         & 24.20 & 0.880 & 0.108 & --   \\
DeepLPF \cite{moran2020deeplpf}     & 24.48 & 0.887 & 0.103 & 7.77 \\
FlexiCurve \cite{li2023flexicurve}   & 24.74 & \underline{0.920} & \underline{0.060} & --   \\
NamedCurves \cite{serrano2024namedcurves}  & 25.20 & 0.906 & \textbf{0.047} & 7.58 \\
BasicEnhancer \cite{song2021starenhancer} & \underline{25.47} & 0.889 & \underline{0.060} & \underline{7.42} \\
\textbf{CA-ATP} & \textbf{26.23} & \textbf{0.925} & \textbf{0.047} & \textbf{6.81} \\
\bottomrule
\end{tabular}
\end{table}

\begin{table*}[!h]
\caption{
Peer comparison on the MIT5K-Star dataset with subsets retouched by five experts (A, B, C, D, and E). The best result is indicated in \textbf{bold}, while the second best is \underline{underlined}. All models are trained once and tested on all five styles. The symbol $^{\dagger}$ indicates models evaluated using attribute vectors, with attribute evaluation following the approach of DiffRetouch \cite{duan2025diffretouch}.
}
\label{table2}
\centering
\renewcommand{\arraystretch}{1.2}
\fontsize{8.5pt}{8.5pt}\selectfont%
\begin{tabular}{ccccccc}
\toprule
\multirow{2}{*}{\textbf{Method}} & \multicolumn{6}{c}{\textbf{PSNR$\uparrow$ / SSIM$\uparrow$}} \\
\cline{2-7}\noalign{\smallskip}
 & \textbf{A} & \textbf{B} & \textbf{C} & \textbf{D} & \textbf{E} & \textbf{Average} \\
\midrule
PIENet \cite{kim2020pienet}  & \underline{21.54}/0.882 & \underline{26.02}/0.948 & 25.29/0.919 & 22.95/0.905 & 24.22/0.925 & 24.00/0.916 \\

TSFlow \cite{kim2023learning} & 20.65/0.869 & 25.34/{0.952} & 25.57/0.935 & 22.48/0.913 & 23.65/0.935 & 23.54/0.921 \\

StarEnhancer \cite{song2021starenhancer}  & 20.75/0.880 & 25.84/{0.952} & \underline{25.73}/0.937 & \underline{23.50/0.922} & \underline{24.60}/0.947 & \underline{24.09}/0.928 \\

DiffRetouch \cite{duan2025diffretouch}  & \textbf{22.76/0.921} & 25.31/\underline{0.960} & 25.33/\underline{0.940} & 22.50/\underline{0.922} & 24.25/\underline{0.952} & 24.03/\underline{0.939} \\

\textbf{CA-ATP} & 21.35/\underline{0.899} & \textbf{27.08/0.976} & \textbf{26.03/0.954} & \textbf{23.85/0.938} & \textbf{25.20/0.961} & \textbf{24.70/0.946} \\
\midrule
DiffRetouch$^{\dagger}$ \cite{duan2025diffretouch}  & 24.48/0.936 & 26.12/0.958 & 26.21/0.944 & 24.51/0.940 & 24.67/0.953 & 25.20/0.946 \\
\textbf{CA-ATP}$^{\dagger}$  & \textbf{25.38/0.946} & \textbf{27.28/0.972} & \textbf{26.77/0.956} & \textbf{26.10/0.958} & \textbf{25.91/0.963} & \textbf{26.29/0.959} \\
\bottomrule
\end{tabular}
\end{table*}

\noindent\textbf{Multi-Style Setup.}
For a fair comparison with existing methods, we follow the setup of StarEnhancer \cite{song2021starenhancer}, resizing all images so that their shorter side is 340 pixels (hereafter referred to as MIT5K-Star). Images from all five styles are divided into training and testing sets. For the PPR10K dataset, we employ the official 360p images, excluding the additional augmentation dataset and portrait masks. The data split follows DiffRetouch \cite{duan2025diffretouch}, with 1,356 groups (8,875 images) for training and 325 groups (2,286 images) for testing. The evaluation metrics for multi-style image retouching are PSNR and SSIM, which are calculated in the same way as DiffRetouch.

\noindent\textbf{Implementation Details.}
All experiments are executed on a single NVIDIA RTX 3090 GPU using PyTorch. The parameters of our model are optimized using the Adam optimizer \cite{kingma2015adam}, with the pretrained CLIP encoder fine-tuned at a learning rate of $1 \times 10^{-5}$, while the remaining modules are trained at a learning rate of $1 \times 10^{-4}$. The number of training epochs is 100, the batch size is 3, and a cosine annealing learning rate schedule \cite{loshchilov2017sgdr} is applied. Besides, we conduct a data augmentation that includes random horizontal and vertical flips, random cropping, and random transposition.  The total loss is a weighted sum of the MSE and SSIM loss functions, with their respective weights $\alpha$ and $\beta$ set to 1.0 and 0.4.

\subsection{Peer Comparison}

\noindent\textbf{Single-Style Methods.}
We first compare our method with several state-of-the-art single-style image retouching methods. The compared methods include UPE \cite{wang2019underexposed} and DeepLPF \cite{moran2020deeplpf}, as well as the curve-based methods (CURL \cite{moran2021curl}, BasicEnhancer \cite{song2021starenhancer}, FlexiCurve \cite{li2023flexicurve} and NamedCurves \cite{serrano2024namedcurves}). All methods are trained and tested according to the single-style setup. The quantitative results are summarized in Table~\ref{table1}. Specifically, our approach achieves the highest PSNR of 26.23 dB and SSIM of 0.925, the lowest color difference with a $\Delta E_{ab}$ of 6.81, and matches the best reported LPIPS score of 0.047 by NamedCurves, which demonstrates the superiority of our method in single-style image retouching for general users. While our method is designed for multi-style image retouching, ‌the integration of a content-adaptive strategy and attribute text representation also achieves significant performance in single-style image retouching for general users. 

\begin{table}[!h]
\caption{Peer comparison on the PPR10K dataset with subsets retouched by three experts. ${\S}$ indicates the single-style method that trains three separate models for the three experts, respectively. The best result is indicated in \textbf{bold}, while the second best is \underline{underlined}. 
}
\label{table3}
\renewcommand{\arraystretch}{1.2}
\fontsize{8.5pt}{8.5pt}\selectfont%
\centering
\begin{tabular}{lccccc}
\toprule
\textbf{Expert} & \textbf{Method} & \textbf{PSNR}$\uparrow$ & \textbf{SSIM}$\uparrow$ & \textbf{LPIPS}$\downarrow$ \\
\midrule
\multirow{5}{*}{A}  &HDRNet$^{\S}$ \cite{gharbi2017deep}  & 23.01 & 0.953 & 0.057 \\
& CSRNet$^{\S}$ \cite{he2020conditional}  & 23.86 & 0.952 & 0.055 \\
& 3DLUT$^{\S}$ \cite{zeng2020learning} & 25.98 & 0.967 & \underline{0.040} \\
& DiffRetouch$^{\dagger}$ \cite{duan2025diffretouch} & \underline{26.23} & \underline{0.970} & \underline{0.040} \\
& \textbf{CA-ATP} & \textbf{26.58} & \textbf{0.975} & \textbf{0.033} \\
\midrule
\multirow{5}{*}{B}  & HDRNet$^{\S}$ \cite{gharbi2017deep} & 23.17 & 0.952 & 0.058 \\
& CSRNet$^{\S}$ \cite{he2020conditional}  & 23.70 & 0.952 & 0.057 \\
& 3DLUT$^{\S}$ \cite{zeng2020learning} & 25.06 & 0.959 & 0.046 \\
& DiffRetouch$^{\dagger}$ \cite{duan2025diffretouch}  & \underline{25.65} & \underline{0.969} & \underline{0.042} \\
& \textbf{CA-ATP}  & \textbf{25.67} & \textbf{0.970} & \textbf{0.035} \\
\midrule
\multirow{5}{*}{C}  &HDRNet$^{\S}$ \cite{gharbi2017deep} & 23.34 & 0.951 & 0.058 \\
& CSRNet$^{\S}$ \cite{he2020conditional} & 23.87 & 0.953 & 0.055 \\
& 3DLUT$^{\S}$ \cite{zeng2020learning} & 25.45 & 0.961 & 0.045 \\
& DiffRetouch$^{\dagger}$ \cite{duan2025diffretouch} & \underline{25.67} & \underline{0.966} & \underline{0.043} \\
& \textbf{CA-ATP}  & \textbf{26.21} & \textbf{0.971} & \textbf{0.034} \\
\bottomrule
\end{tabular}

\end{table}

\noindent\textbf{Multi-Style Methods.}
For multi-style image retouching, we compare our method with some advanced approaches, including PIENet \cite{kim2020pienet}, StarEnhancer \cite{song2021starenhancer}, TSFlow \cite{kim2023learning}, and DiffRetouch \cite{duan2025diffretouch}. These methods typically employ a single model to learn diverse retouching preferences, which is different from single-style methods \cite{wang2019underexposed} that can only learn one style preference in a single model. A critical distinction among these methods lies in their inference protocols. While PIENet and StarEnhancer rely on deep networks for abstract feature representation, DiffRetouch and our model employ style vectors for feature representation. However, the inference protocol of DiffRetouch requires obtaining these vectors from paired target images, making this approach impractical in real-world scenarios where target images are unavailable. 

To ensure a fair comparison in a practical condition, we limit DiffRetouch’s dependence on target images during the inference phase. Instead of accessing individual target images during inference, we compute the average attribute vectors derived from known retouched image pairs for each user and employ these fixed user-specific vectors to guide the inference process. All attribute vectors for averaging and evaluation are provided by DiffRetouch. In contrast, our proposed CA-ATP inherently eliminates the need for target images by deriving target attribute vectors through the trained ATP module. The quantitative results on the MIT5K-Star dataset \cite{song2021starenhancer} are presented in Table~\ref{table2}. Our method demonstrates superior performance across both experimental settings. In the practical setting, compared to PIENet, StarEnhancer, TSFlow, and DiffRetouch, our CA-ATP achieves the highest average PSNR of 24.70 dB and SSIM of 0.946. Notably, when DiffRetouch is adapted to employ averaged attribute vectors, our approach outperforms it by 0.67 dB in PSNR.

Furthermore, we employ an inference protocol based on target image attributes (denoted by ${\dagger}$) to validate our model, enabling direct comparison with the original DiffRetouch implementation. Under this setting, our method (CA-ATP$^{\dagger}$) achieves significantly superior performance relative to DiffRetouch$^{\dagger}$, attaining an average PSNR of 26.29 dB compared to 25.20 dB, and an average SSIM of 0.959 versus 0.946. These results confirm the superiority of the proposed approach in learning multi-style image retouching based on user-defined style preferences.

\begin{figure*}[!ht]
\centering
\includegraphics[width=\textwidth]{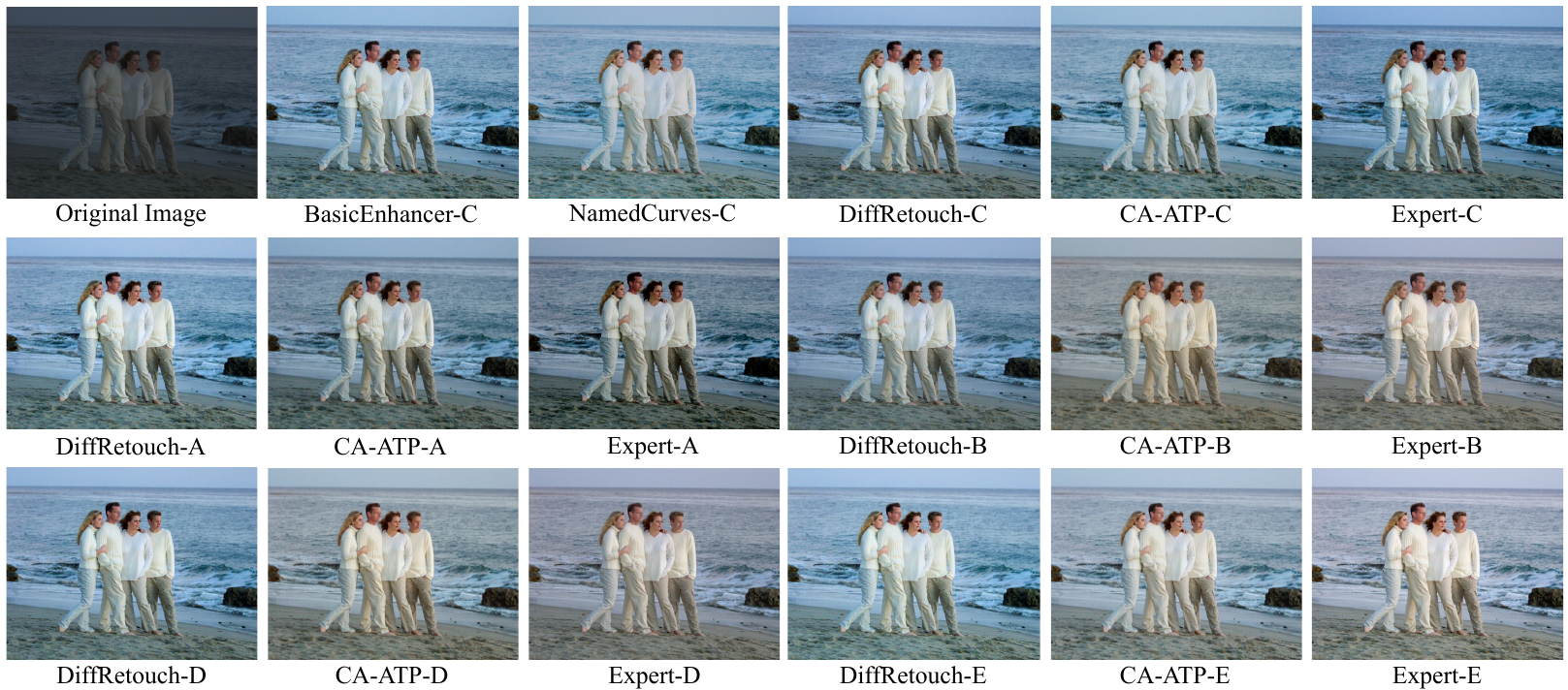}
\caption{Visual comparison on the MIT5K dataset with an image retouched by five experts (A, B, C, D, and E). For example, ``-C'' denotes the results retouched by expert C and different retouching methods. Since BasicEnhancer and NamedCurves are unable to produce multiple retouching styles, only the results corresponding to Expert-C are shown.}
\label{fig4}
\end{figure*}

To visually demonstrate the retouching performance of our model, we conduct a visual analysis of our method and several typical methods (including the single-style BasicEnhancer, NamedCurves, and the multi-style DiffRetouch) on a randomly selected image and its variants from the MIT5K dataset, as shown in Fig.~\ref{fig4}. The first row shows single-style retouching results, while the second and third rows present multi-style comparisons. ‌Compared to other methods, our approach consistently achieves higher visual alignment with expert retouching in both single-style and multi-style scenarios. This indicates that the proposed CA-ATP effectively captures style preferences, delivering high-quality retouching images for both general and diverse individual users.

To further verify the generalization of our method, we also conduct experiments on the PPR10K dataset \cite{liang2021ppr10k}, where each image is retouched by three different experts. We compare our method with several representative methods. As listed in Table~\ref{table3}, our method not only outperforms single-style methods (HDRNet \cite{gharbi2017deep}, CSRNet \cite{he2020conditional}, and 3DLUT \cite{zeng2020learning}) but also achieves superior results compared to DiffRetouch$^{\dagger}$. It is noteworthy that even without accessing the attribute vectors of the target images, our CA-ATP model surpasses the performance of DiffRetouch$^{\dagger}$, which explicitly depends on user style vectors extracted from target images during inference. Specifically, our CA-ATP model achieves the best performance across all three expert preferences, yielding the highest PSNR and SSIM values as well as the lowest LPIPS score. Through the aforementioned peer comparison, we can conclude that our proposed content-adaptive image retouching method based on attribute-text representation is effective in modeling user-defined style preferences.

\subsection{Ablation Study}
\label{sec:ablation}

\begin{table}[h]
\caption{
Ablation study results of our model on the MIT5K-UPE dataset. The best result is indicated in \textbf{bold}, while the second best is \underline{underlined}. 
}
\label{table4}
\renewcommand{\arraystretch}{1.2}
\fontsize{8.5pt}{8.5pt}\selectfont%
\centering
\begin{tabular}{lcc}
\toprule
\textbf{method} & \textbf{PSNR}$\uparrow$ & \textbf{SSIM}$\uparrow$ \\
\midrule
CA-ATP w/o U-Net & 25.81 & 0.916 \\
CA-ATP w/o ATP & \underline{26.10} & \underline{0.923} \\
\textbf{CA-ATP} & \textbf{26.23} & \textbf{0.925} \\
\bottomrule
\end{tabular}
\end{table}

To verify the contributions of the ATP module and the content‑adaptive U‑Net, ablation results are reported in Table~\ref{table4}. Specifically, ``CA-ATP w/o ATP'' denotes our model variant without the ATP module, which only employs image modalities for feature representation. ``CA-ATP w/o U-Net'' indicates our model variant excluding the U-Net architecture, utilizing a monolithic curve mapping.
Compared with CA‑ATP w/o ATP, incorporating attribute text descriptions slightly improves PSNR (from 26.10 dB to 26.23 dB) and SSIM (from 0.923 to 0.925). 
Such a marginal performance improvement is anticipated, given that this ablation study is performed under a single fixed style (Expert C). The principal function of the ATP module is not to maximize fidelity under a constrained stylistic condition, but rather to explicitly enable style-controllable and user-friendly image retouching based on attribute text representations. 
Second, the integration of the content‑adaptive U‑Net module yields a clear performance gain (PSNR improves from 25.81 dB to 26.23 dB, and SSIM rises from 0.916 to 0.925), which validates the effectiveness of the content‑adaptive color mapping strategy.

\begin{table}[h]
\caption{Quantitative comparison of color diversity on the MIT5K-UPE dataset, which is measured by the average unique color count (distinct RGB values) for different conditions.}
\label{table_unique_color}
\renewcommand{\arraystretch}{1.2}
\fontsize{8.5pt}{8.5pt}\selectfont%
\centering
\begin{tabular}{lcccc}
\toprule
\textbf{Condition} & \textbf{Original} & \textbf{w/o U-Net} & \textbf{Target} & \textbf{CA-ATP} \\
\midrule
Unique color count & 19835 & 11003 & 68139 & 70452\\
\bottomrule
\end{tabular}
\end{table}

To validate the necessity of content-adaptive color mapping enabled by the U-Net architecture, we conduct an ablation study using the unique color count (the number of distinct RGB values) as a metric for color diversity. The quantitative results, calculated as the average across the MIT5K-UPE test set, are summarized in Table~\ref{table_unique_color}. The original images exhibit an average of 19,835 unique colors. When the content-adaptive U-Net module ($\text{CA-ATP w/o U-Net} $) is removed from our model, the unique color count declines markedly to 11,003. This indicates that merely using monotonic color mapping cannot learn diverse color transformations. In contrast, the complete $\text{CA-ATP}$ model utilizing the content-adaptive U-Net achieves a significantly higher count of 70,452, which closely aligns with the statistical average of 68,139 unique colors observed in expert‑retouched target images. This finding robustly demonstrates that our content-adaptive framework effectively preserves and enhances color richness by adaptively modulating color curves based on the perceptual content of images, yielding color vibrancy and diversity comparable to expert-retouched images.

\begin{table}[h]
\caption{Ablation study results of the hyperparameter $N$. The best result is indicated in \textbf{bold}, while the second best is \underline{underlined}. As shown in Fig. \ref{fig2}, $N$ is the number of content weight maps in our model.}
\label{table5}
\renewcommand{\arraystretch}{1.2}
\fontsize{8.5pt}{8.5pt}\selectfont%
\centering
\begin{tabular}{lccccc}
\toprule
\textbf{N} & \textbf{1} & \textbf{3} & \textbf{5} & \textbf{7} & \textbf{9} \\
\midrule
PSNR$\uparrow$  & 25.85 & 26.19 & \textbf{26.23} & \underline{26.20} & 26.19 \\
SSIM$\uparrow$  & 0.917 & \underline{0.924} & \textbf{0.925} & 0.923 & \textbf{0.925} \\
\bottomrule
\end{tabular}
\end{table}

\begin{table}[!ht]
\caption{
Ablation study of the U‑Net variants on the MIT5K‑UPE dataset. The best and second‑best results are highlighted in \textbf{bold} and \underline{underline}, respectively.
}
\label{table8}
\centering
\renewcommand{\arraystretch}{1.2}
\fontsize{8.5pt}{8.5pt}\selectfont%
\begin{tabular}{lcc}
\toprule
\textbf{method} & \textbf{PSNR}$\uparrow$ & \textbf{SSIM}$\uparrow$ \\
\midrule
Baseline & 25.75 & 0.917 \\
CNN Encoder & 25.60 & 0.914 \\
CNN U-Net & \underline{26.06} & \underline{0.923} \\
Transformer Encoder & \underline{26.06} & 0.917 \\
Transformer U-Net (Ours) & \textbf{26.23} & \textbf{0.925} \\
\bottomrule
\end{tabular}
\end{table}

To evaluate the effect of hyperparameter $N$, an ablation study is conducted by varying the value of $N$ and evaluating the model on the MIT5K-UPE dataset. The corresponding test results are summarized in Table~\ref{table5}. As shown in the table, model performance improves consistently as $N$ increases from 1 to 5, indicating that greater diversity in content adaptation enhances retouching effectiveness. However, when $N > 5$, the performance gain of our model saturates. This is because the multimodal feature $\bm{f}_i$ must generate parameters for all $N$ basis curves during the curve generation process. When $N$ becomes excessively large, the representational capacity of the feature saturates, limiting its ability to produce additional curves that contribute meaningfully to performance. Consequently, excessive values of $N$ introduce unnecessary computational overhead without yielding further effective information. Based on these findings, $N=5$ is selected as the optimal trade-off between performance and computational efficiency.

To verify the effectiveness of U-Net in producing content-aware weight maps, we further conduct an ablation study on the MIT5K-UPE dataset \cite{wang2019underexposed}.  Specifically, ``Baseline'' denotes our model without the proposed U‑Net. ``CNN U‑Net'' and ``Transformer U‑Net'' use full U‑Net architectures composed of CNN and Transformer blocks, respectively. ``CNN Encoder'' and ``Transformer Encoder'' are encoder‑only variants, excluding both the decoder and skip connections. Table~\ref{table8} lists the results of different U-Net architectures. First, the baseline without this module achieves 25.75 dB PSNR and 0.917 SSIM. Incorporating a CNN-based U-Net (CNN U-Net) increases PSNR to 26.06 dB and SSIM to 0.923, demonstrating the benefit of our content-aware module. Then, we compare the proposed U-Net with its encoder-only variants that remove the decoder and skip connections. The CNN encoder yields a PSNR of 25.60 dB and an SSIM of 0.914, which are lower than the baseline, respectively. The Transformer encoder (26.06 dB / 0.917) outperforms the CNN encoder, but still falls short of the full U-Net. These results highlight the importance of skip connections in U-Net and multi-scale fusion for integrating the content-adaptive curve mapping module.

Finally, we examine the effect of the building blocks within the U-Net architecture by comparing the CNN-based U-Net (CNN U-Net) with our Transformer-based U-Net (Transformer U-Net). The Transformer U-Net achieves the best performance. This improvement can be attributed to the superior long-range dependency modeling of the Transformer block, which enables more effective utilization of rich contextual information in color space and produces more accurate content-aware retouching results.

\subsection{Visual Analysis}

\begin{figure*}[!ht]
\centering
\includegraphics[width=\textwidth]{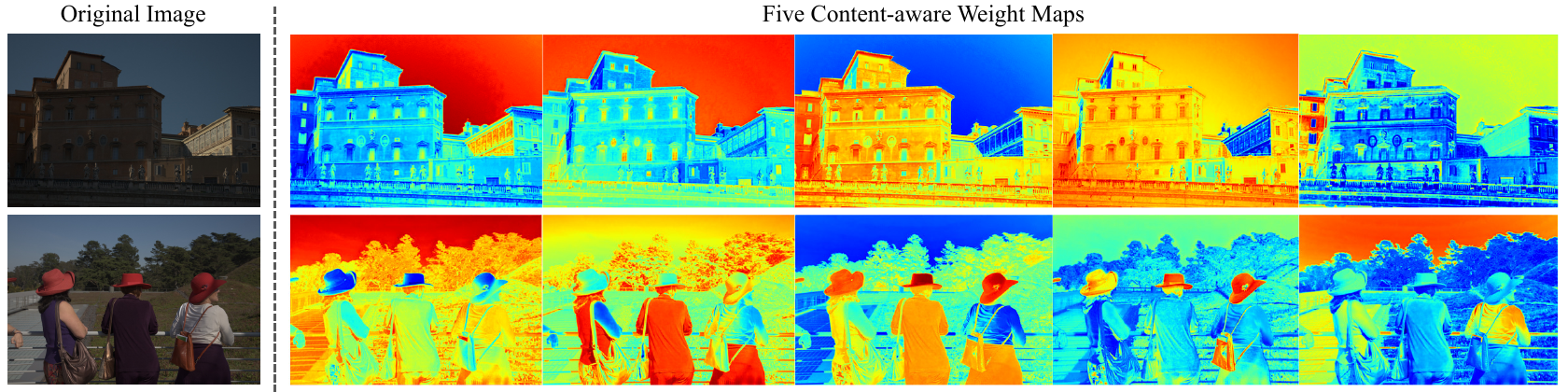}
\caption{Visualization of content-aware weight maps. In each row, the first image shows an original image, and the subsequent five images show different weight maps used for color mapping. Red pixels denote regions with higher weights, while blue pixels represent lower ones.}
\label{fig5}
\end{figure*}

To intuitively demonstrate the effectiveness of content-aware weight maps, we visualize two randomly selected original images from the MIT5K dataset~\cite{bychkovsky2011learning} and their five corresponding content-aware weight maps in Fig.~\ref{fig5}. For the architectural scene in the first row, the learned maps exhibit a strong capability in structural segmentation, where the sky, the building facade, and the shadow and highlight regions are captured by different weight maps, respectively. Similarly, in the second row featuring human subjects, the weight maps show clear content awareness by distinctively separating the foreground figures (e.g., hats and clothing) from the background vegetation and sky. These visualizations substantiate that our CA-ATP model can spatially disentangle complex image content into distinct regions. This mechanism facilitates the precise application of diverse color mapping strategies to different content areas, allowing the retouched images to be more consistent with the human-retouched results.

\begin{figure*}[!ht]
\centering
\includegraphics[width=0.95\textwidth]{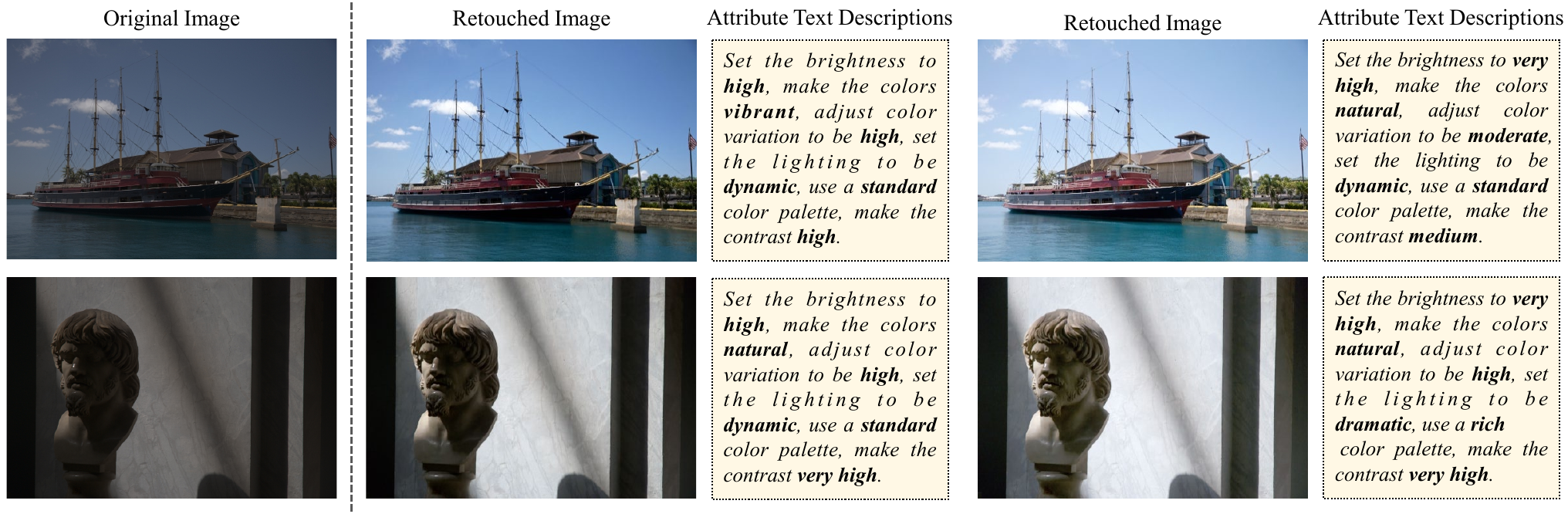}
\caption{Visual comparison of text-guided image retouching. The figure presents two examples (top: a ship, bottom: a statue). The first column shows the original image. For each original image, the two subsequent columns display different retouched image results retouched by the corresponding attribute text descriptions on the right.}
\label{fig6}
\end{figure*}

To further demonstrate the intuitiveness and effectiveness of the proposed attribute-based text in guiding multi-style image retouching, Fig.~\ref{fig6} presents two randomly selected original images from the MIT5K dataset~\cite{bychkovsky2011learning} alongside two retouched results for each, reflecting user-defined style preferences. The first row (ship example) illustrates how different attribute descriptions guide text-driven retouching across distinct versions. Both outputs show increased brightness consistent with their prompts, yet they exhibit qualitative differences in saturation and intensity attributable to the specific text inputs. Similarly, the second row (statue example) demonstrates the handling of low-light images with deep shadows. While the first result enhances brightness and contrast to accentuate shadow lines, the second yields a notably brighter output, demonstrating our model’s ability to follow differing brightness constraints described in the text. Therefore, the proposed attribute-based text representation not only intuitively characterizes diverse user style preferences but also achieves semantically consistent retouching results that align with the attribute-based text descriptions, achieving user-friendly guidance for image retouching.

\section{Conclusion}
In this paper, we have proposed a content-adaptive image retouching method guided by attribute-based text representation. To address the limitation of color diversity in existing color mapping methods, we develop a content-aware adaptation strategy. Extensive experiments demonstrate that our model can dynamically capture color diversity according to image content, effectively transcending conventional color mapping constraints. Furthermore, we have proposed an effective method for characterizing user-defined style preferences, which employs attribute text descriptions to intuitively specify desired retouching direction and intensity. This offers a user-friendly and intuitively controllable solution for multi-style image retouching. By introducing these two strategies, the proposed CA-ATP model significantly outperforms state-of-the-art approaches in comprehensive experiments. In color mapping-based image retouching tasks, our CA-ATP model can effectively learn rich color mapping relationships and yield retouching results that are highly consistent with user preferences.

\bibliographystyle{IEEEtran}
\bibliography{main}

\vfill

\end{document}